\documentclass[times,10pt,twocolumn]{article}
\usepackage{latex8}
\usepackage{times}
\usepackage{balance}
\usepackage{url}
\AtBeginDocument{%
  \providecommand\BibTeX{{%
    \normalfont B\kern-0.5em{\scshape i\kern-0.25em b}\kern-0.8em\TeX}}}

\usepackage{xspace}
\usepackage{caption}
\usepackage[skip=0cm,list=true,labelfont=bf]{subcaption}

\usepackage{soul}
\usepackage{lipsum,graphicx,multicol}
\usepackage{tabularx}
\newcolumntype{Y}{>{\centering\arraybackslash}X}

\usepackage{color, colortbl}
\definecolor{Gray}{gray}{0.9}
\usepackage[first=0,last=9]{lcg}

\usepackage[]{graphicx} % omit 'demo' for real document

\usepackage{algorithm}
\usepackage[noend]{algpseudocode}
\usepackage{nicefrac}
\providecommand{\keywords}[1]{\textbf{\textit{Index terms---}} #1}

\usepackage{amsmath}
\usepackage{algorithm}
\usepackage[noend]{algpseudocode}

\makeatletter
\def\BState{\State\hskip-\ALG@thistlm}
\makeatother

\begin{document}

\title{Benchmarking of EEG Analysis Techniques for Parkinson's Disease Diagnosis: A Comparison between Traditional ML Methods and Foundation DL Methods}
\author{Danilo Avola\textsuperscript{1}, 
Andrea Bernardini\textsuperscript{2}, 
Giancarlo Crocetti\textsuperscript{3}, 
Andrea Ladogana\textsuperscript{1}, \\
Mario Lezoche\textsuperscript{4},
Maurizio Mancini\textsuperscript{1}, 
Daniele Pannone\textsuperscript{1}, and Amedeo Ranaldi\textsuperscript{1}\\
\textsuperscript{1}Department of Computer Science, Sapienza University of Rome\\
Via Salaria 113, 00198, Rome (RM), Italy\\
\{avola,danese,m.mancini,pannone,ranaldi\}@di.uniroma1.it\\
\textsuperscript{2}Fondazione Ugo Bordoni\\
Viale del Policlinico 147, 00161, Rome (RM), Italy\\
abernardini@fub.it\\
\textsuperscript{3}Department of Computer Science and Mathematics, \\
The L. H. and W. L. Collins College of Professional Studies, \\ 
St. John's University, USA\\
crocettg@stjohns.edu \\
\textsuperscript{3}University of Lorraine, CNRS, CRAN\\
Nancy, F-54000, France\\
mario.lezoche@univ-lorraine.fr
}

\maketitle
\thispagestyle{empty}

\begin{abstract}
Parkinson’s Disease (PD) is a progressive neurodegenerative disorder that affects motor and cognitive functions, with early diagnosis being critical for effective clinical intervention. Electroencephalography (EEG) offers a non-invasive and cost-effective means of detecting PD-related neural alterations, yet the development of reliable automated diagnostic models remains a challenge. In this study, we conduct a systematic benchmark of traditional machine learning (ML) and deep learning (DL) models for classifying PD using a publicly available oddball task dataset. Our aim is to lay the groundwork for developing an effective learning system and to determine which approach produces the best results. We implement a unified seven-step preprocessing pipeline and apply consistent subject-wise cross-validation and evaluation criteria to ensure comparability across models. Our results demonstrate that while baseline deep learning architectures, particularly CNN–LSTM models, achieve the best performance compared to other deep learning architectures, underlining the importance of capturing long-range temporal dependencies, several traditional classifiers such as XGBoost also offer strong predictive accuracy and calibrated decision boundaries. By rigorously comparing these baselines, our work provides a solid reference framework for future studies aiming to develop and evaluate more complex or specialized architectures. Establishing a reliable set of baseline results is essential to contextualize improvements introduced by novel methods, ensuring scientific rigor and reproducibility in the evolving field of EEG-based neurodiagnostics.
\end{abstract}


% \maketitle
\section{Introduction}
PD is the second most common neurodegenerative disorder globally, affecting over 10 million people, and is characterized by motor symptoms such as bradykinesia, rigidity, and tremor \cite{kalia2015parkinsons}. Timely and accurate detection of PD is crucial for effective clinical management, yet current diagnostic approaches heavily rely on subjective assessments or expensive imaging methods. EEG, however, offers a non-invasive, economical, and scalable alternative, capturing distinctive PD-related oscillatory slowing even in non-demented individuals \cite{stoffers2007slowing}.

In recent years, ML and DL techniques have been increasingly applied to EEG data for PD diagnosis, yielding promising results. These approaches range from classical ML models that rely on handcrafted spectral or statistical features, such as support vector machines (SVM) \cite{ds004574, zhang2018eegpd, zhang2021} and random forests \cite{liu2019random, LEE2021109282, CHATURVEDI20191937, GERAEDTS20211041}, to complex deep architectures methods including convolutional neural networks (CNN) and recurrent neural networks (RNN) trained end-to-end \cite{SHAH202075, 10.3389, 10.1007}. 

Despite this progress, much of the existing literature lacks methodological consistency: models are evaluated under different preprocessing strategies, data splits, and performance metrics, often on limited or non-public datasets. This heterogeneity undermines reproducibility and makes it difficult to assess whether improvements reported in complex architectures reflect genuine gains or are confounded by uncontrolled variables.

To address this gap, the present work provides a rigorous and systematic comparison of baseline ML and DL models on a publicly available EEG dataset recorded during an oddball task, a paradigm known to evoke cognitive responses altered in PD. Rather than pursuing novel or highly specialized architectures, our goal is to establish a transparent and reproducible benchmark composed of representative baseline models from both traditional and deep learning paradigms. These include classical classifiers such as logistic regression, support vector machines, and XGBoost, as well as foundational DL architectures like CNNs, Long short-term memory (LSTM), Temporal Convolutional networks (TCNs), and CNN–LSTM hybrids.

By implementing a unified seven-stage preprocessing pipeline, strict subject-wise train/test splits, and a standardized evaluation framework, we aim to isolate model-specific behavior from dataset or pipeline variability. This controlled benchmarking effort lays a critical foundation for future research, offering reference points against which new or more complex models can be meaningfully assessed. Establishing strong and interpretable baselines is not only essential for scientific rigor but also provides the clarity needed to advance the field of EEG-based PD diagnostics in a reproducible and clinically relevant direction. In summary, the main contributions of the paper are as follows:

\begin{itemize}
\item We present a comprehensive benchmark of both traditional machine learning and deep learning baseline models for Parkinson’s disease classification using EEG data from a publicly available oddball task dataset;

\item We introduce a unified seven-step preprocessing pipeline, covering filtering, artifact correction, epoching, data augmentation, spectral decomposition, normalization, and reshaping, ensuring consistency across all models and reproducibility for future research;

\item We evaluate five traditional classifiers (logistic regression, SVM, random forest, $k$-NN, and XGBoost) and six deep learning models (CNN, CNN-GAP, LSTM, two CNN–LSTM hybrids, and TCN), comparing performance using accuracy, precision, recall, F1-score, AUC–ROC, and log loss;

\item We demonstrate that baseline hybrid CNN–LSTM architectures outperform other deep learning approaches, highlighting the importance of modeling both spatial and temporal structure in EEG data. Furthermore, we show that XGBoost achieves remarkably strong performance on EEG-based PD classification, outperforming most traditional classifiers and even rivaling deep learning models in terms of accuracy, AUC–ROC, and reliability of predictions;

\item We provide a clear and interpretable performance baseline that serves as a foundation for the development and evaluation of more complex or domain-specific models in future EEG-based diagnostic studies.
\end{itemize}

The remainder of the paper is structured as follows: Section~\ref{sec:materials_and_methods} details the dataset and methodological framework; Section~\ref{sec:results} presents the experimental results; and Section~\ref{sec:conclusion} discusses our findings and concludes with key takeaways and directions for future work.

\section{Related Work} \label{sec:related_work}
Numerous neurophysiological studies have documented distinct alterations in resting-state oscillatory activity associated with Parkinson’s Disease (PD). Early EEG research on non-demented PD patients consistently highlighted occipital rhythm slowing and reduced beta-band power compared to healthy individuals \cite{soikkeli1991slowing, neufeld1988eeg}. More recent studies employing magnetoencephalography (MEG) further validated these findings, demonstrating increased theta and low-alpha activity coupled with reduced beta/gamma power even in untreated (de novo) patients. These observations suggest oscillatory slowing is a fundamental characteristic of early-stage PD, largely unaffected by dopaminergic therapy \cite{bosboom2006meg, stoffers2007slowing}, across patient subgroups examined.

Oscillatory changes have also been linked directly to cognitive and motor dysfunction in PD. Bosboom et al. \cite{bosboom2006meg} demonstrated correlations between elevated theta power in occipital channels and subtle cognitive impairments measured by the CAMCOG scale. Cassidy and Brown \cite{cassidy2001eeg} provided evidence that levodopa treatment modifies cortico-cortical connectivity, mitigating abnormal synchronization patterns within frontal networks. Similarly, Kotini et al. \cite{kotini2005meg} found associations between alterations in sensorimotor alpha rhythms and the severity of motor symptoms (UPDRS-III), indicating interactions between cortical and subcortical regions via basal ganglia circuits.

From a computational standpoint, classical machine-learning approaches leveraging handcrafted spectral and temporal features have demonstrated effective discrimination between PD and control subjects \cite{CHATURVEDI20191937, GERAEDTS20211041, LEE2021109282, liu2019random, zhang2018eegpd}. Zhang et al. \cite{zhang2018eegpd} achieved an accuracy of 75\% using support vector machines (SVMs) applied to delta/theta power bands. Liu et al. \cite{liu2019random} reported comparable performance utilizing random forest classifiers that integrated spectral and entropy-based features, though on small, limited cohorts. 

Initial attempts to apply deep learning models to the PD task used Multi-Layer Perceptron networks (MLP) \cite{6610487, 7592119, Hussain_2019, Barua_2019, Rahman2020, 9342332}. In \cite{6610487} The authors demonstrated that integrating spatial, spectral, and temporal features of EEG signals, using wavelet coefficients, spatial correlations, and cross-frequency energy ratios, can effectively predict the transition to Freezing of Gait in Parkinson’s disease patients using a three-layer MLP trained using the Levenberg-Marquardt algorithm with improved predictive accuracy.

Advances in deep learning have seen convolutional neural networks (CNNs) and hybrid CNN–LSTM architectures applied directly to raw EEG data or time–frequency representations, achieving optimal results, albeit predominantly using non-public or single-center datasets \cite{a15010005, Shu2020, huang2020cnn, 9430513, 8969309, 9023190, wang2021hybrid}. Emad Arasteh et al. \cite{a15010005} propose a novel method that transforms directional connectivity patterns from resting-state EEG into 2D images, which are then classified using a fine-tuned VGG-16 deep transfer learning model. In \cite{9430513} the authors proposed a CNN-based model that uses smoothed pseudo-Wigner Ville distribution (SPWVD) to convert EEG signals into time–frequency images.

Despite significant progress, existing literature remains fragmented by heterogeneity in preprocessing methods, diverse feature extraction approaches, and inconsistent validation techniques, complicating meaningful comparisons across studies. Our study addresses these gaps by systematically evaluating a wide array of baseline ML and DL techniques within a unified methodological framework, using a publicly accessible EEG dataset. This comprehensive benchmarking effort provides clarity regarding the relative merits of handcrafted feature extraction versus deep representation learning in detecting Parkinson’s Disease.

\section{Materials and Methods} \label{sec:materials_and_methods}
In this section, we describe the dataset, the preprocessing pipeline, and the benchmarking methodology used to evaluate traditional ML and DL models.

\subsection{Dataset}

The dataset selected for this study is the Iowa dataset for the Cross-modal Oddball Task \cite{ds004574}. This dataset, utilized in Singh et al.’s study, was collected from patients and controls recruited through the University of Iowa Movement Disorders Clinic, a leading academic center with a strong clinical and research infrastructure.  Among the cognitive paradigms employed in the Singh et al. dataset, the oddball task emerges as the most practical and diagnostically informative single-task paradigm for evaluating cognitive dysfunction in Parkinson’s disease (PD). While all three tasks engage executive functions, the oddball task offers several methodological and neurophysiological advantages that justify its isolated use in future research or clinical applications. The oddball task reliably elicited midfrontal delta and theta oscillations in response to novel auditory stimuli, and these oscillations were significantly attenuated in PD patients with cognitive impairment, particularly those with dementia (PDD). These findings suggest that even basic novelty detection mechanisms, as probed by the oddball task, are degraded in PD-related cognitive decline, and that delta/theta activity during this task is a sensitive marker of such degradation. This concept suits perfectly for our task.

This dataset included 146 participants (98 with Parkinson’s disease and 48 controls) tested between 2017 and 2021. During EEG recording with a 64-channel BrainVision cap, subjects performed an oddball task requiring a directional response to a white arrow. Each trial was preceded by simultaneous visual and auditory pre-cues 500 ms before the arrow. Analyses focused on trials with either standard pre-cues or an auditory oddball.

\begin{figure}[t]
\centering
\includegraphics[width=0.5\linewidth]{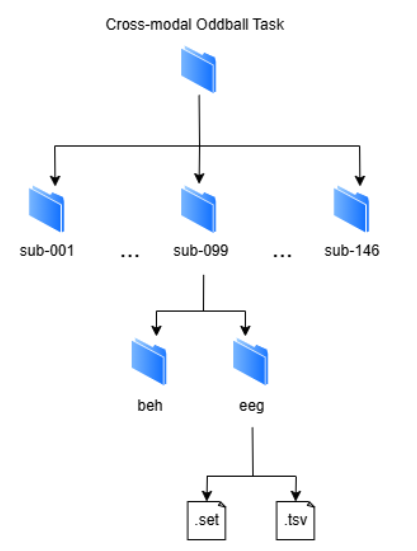}
\caption{Filesystem structure of the dataset.}
\label{fig:datasetTree}
\end{figure}

All raw EEG data were organized under a common \verb|base_path|, with one subdirectory per participant named by their unique ID (Figure \ref{fig:datasetTree}). Each participant folder contained:
\begin{itemize}
  \item \verb|eeg/<ID>_task-Oddball_eeg.set|  
    – continuous EEG recording in EEGLAB format.
  \item \verb|eeg/<ID>_task-Oddball_events.tsv|  
    – tab‐delimited event file listing timestamps and codes for each stimulus.
\end{itemize}
The next section provides details on the preprocessing methods applied to the EEG data.

\subsection{Preprocessing} \label{sec:preprocessing}
\begin{figure}[t]
  \centering
  \includegraphics[width=0.3\linewidth]{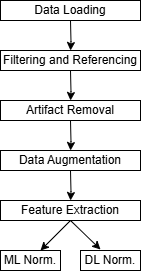}
  \caption{Overview of the EEG preprocessing and feature‐normalization workflow.}

  \label{fig:preproc_flowchart}
\end{figure}

The preprocessing pipeline, essential to ensure data quality and consistency across machine learning and deep learning models, was implemented using MNE and scikit-learn. The raw EEG data, initially stored in EEGLAB \verb|.set| format with accompanying metadata in \verb|.tsv| files, were first imported into the MNE framework for unified handling. An overview of the pipeline is shown in Figure \ref{fig:preproc_flowchart}.

Signal preprocessing began with temporal filtering, where a finite impulse response (FIR) band-pass filter between 1 and 50 Hz was applied to remove low-frequency drift and high-frequency noise. Additionally, a notch filter at 60 Hz was used to suppress power line interference. Channels FT9, FT10, TP9, and TP10 were excluded due to their frequent contamination by movement-related artifacts. Re-referencing was performed using a common average reference approach to improve the spatial uniformity of the signal.

Artifact removal followed a two-step strategy. First, the FASTER algorithm \cite{nolan2010faster} was employed to detect and interpolate noisy channels. Then, Independent Component Analysis (ICA) was used to identify and remove components corresponding to eye movements, blinks, and other stereotypical artifacts. To further clean the data, epochs exceeding 100 $\mu$V in peak-to-peak amplitude or showing joint-probability z-scores above ±3 were rejected.

The continuous EEG signal was segmented into epochs ranging from –1.0 to +2.5 seconds relative to each cue onset. This window was selected to capture both preparatory and response-related neural dynamics. To improve generalization and address class imbalance, data augmentation techniques were applied. Each trial was duplicated with either the addition of low-amplitude Gaussian noise, where the noise standard deviation was set to 1\% of the original trial's standard deviation, or by applying random temporal shifts of up to ±100 milliseconds. This process expanded the number of trials from 98 to 170 for the Parkinson's group and from 48 to 174 for the control group.

Following epoching and augmentation, spectral features were extracted using multitaper power spectral density (PSD) estimation in the 1–50 Hz range. The resulting frequency-domain representations were then aggregated into standard frequency bands,delta, theta, alpha, beta, and gamma, yielding a compact three-dimensional tensor structure defined by the number of bands, time samples, and frequency bins, for efficient analysis.

To prepare the data for model training, normalization and reshaping were tailored to the specific requirements of the ML and DL pipelines. In the ML branch, the tensors were flattened into feature vectors and scaled using a standard normalization fitted exclusively on the training fold to prevent information leakage. In the DL branch, data were formatted into four-dimensional arrays with dimensions corresponding to samples, time, frequency, and channels. Normalization in this case was performed per channel, using either a skewness-aware logarithmic transformation followed by min–max scaling, or direct min–max normalization to the [0, 1] range, depending on the architecture employed.
\subsection{Benchmarking Methodology} \label{sec:benchmarking_methodology}
To assess the efficacy of machine learning and deep learning approaches in classifying Parkinson’s disease based on EEG data, we implemented a comprehensive experimental design encompassing data partitioning, model training, evaluation, and efficiency analysis.

The dataset was divided with a strict subject-wise stratification strategy to ensure that no individual contributed data to both the training and test sets, thereby eliminating any potential data leakage. Seventy percent of the subjects were randomly assigned to the training set, while the remaining thirty percent formed a held-out test set used exclusively for final evaluation. Within the training partition, a five-fold cross-validation scheme was employed to guide hyperparameter selection and model validation. All normalization procedures were applied within each training fold to prevent information from the validation or test sets from influencing model calibration and ensure unbiased evaluation.

In the traditional ML framework, we evaluated five classifiers. The classifiers included logistic regression with L2 regularization (solver set to 'liblinear' and penalty parameter C = 1.0), a support vector machine with a radial basis function kernel ($C = 10$, $\gamma = 0.1$), a random forest consisting of 100 decision trees with a maximum depth of 10, a $k$-NN model using k = 5 with distance-based weighting, and a gradient boosting classifier via XGBoost with 100 estimators, maximum tree depth of 6, and a learning rate of 0.1. For each of these models, hyperparameters were optimized through a grid search performed independently within each fold of the cross-validation procedure, ensuring an unbiased selection process and robust model performance overall.

For the DL-based models, we employed TensorFlow to construct and train six baseline architectures: CNN, CNN with Global Average Pooling (CNN-GAP), LSTM, two versions of hybrid CNN–LSTM model, and a TCN model. These baseline models are accurately selected to explore the possible ways this kind of task can be addressed. EEG signals are structured across time, frequency, and channel dimensions. CNNs are well-suited for capturing local patterns and spatial dependencies in structured input like time–frequency maps or multichannel EEG matrices. In the case of Parkinson’s disease, disease-related signatures may manifest as localized abnormalities in spectral power (e.g., increased beta-band activity) or altered synchronization patterns. CNNs are capable of learning these local representations without the need for hand-crafted features. Furthermore, EEG signals are subject to inter-subject variability and noise. Convolutional filters can learn to generalize across different subjects and sessions by focusing on repeatable frequency–channel patterns, helping the model become more robust to individual differences and recording noise.

With CNN-GAP we explore the advantages of Global Average Pooling. GAP reduces the dimensionality of feature maps while preserving the mean activation over regions. In EEG, this can help retain the global structure of the signal, particularly relevant when pathological activity is diffuse or distributed across brain regions, as in PD. Instead of relying on specific localized activations (as with max pooling or flattening), GAP averages over the entire spatial or temporal dimension of the final feature map. This is ideal when the discriminative information is globally distributed, for example, in PD-related EEG changes that occur across widespread brain regions rather than in isolated areas. GAP thus enables the model to focus on overall activity patterns rather than precise spatial arrangements, which may vary across subjects. Furthermore, EEG data varies significantly from person to person due to anatomy, electrode placement, and noise. GAP offers robustness by smoothing out local idiosyncrasies and emphasizing consistently recurring global patterns, which are more likely to reflect genuine neurophysiological markers of disease rather than individual variability.

The second group of baseline models is selected to explore the temporal information contained in the EEG. Parkinson’s disease is not only characterized by static spectral abnormalities but also by altered temporal dynamics, such as changes in neural oscillation stability, synchronization, and phase relationships. LSTM networks are explicitly designed to capture long-range dependencies and temporal sequences, making them ideal for modeling the evolution of EEG patterns. LSTMs can naturally handle temporal variability and noise by learning to prioritize consistent signal dynamics over time. This is valuable in EEG data, where noise and transient artifacts can introduce short-term fluctuations that may not be relevant to the disease state.

In the literature is also very often used a combination of LSTM and CNN. CNNs excel at extracting spatial and frequency-domain patterns from short segments of EEG, such as detecting bursts of synchrony or localized rhythms. However, they do not model how these patterns evolve over time. By using CNN layers first, the model can extract rich local features, which are then passed to LSTM layers to capture temporal dependencies between these learned patterns.
This is particularly useful for EEG, where pathological changes often manifest as both localized spectral features and slower dynamics. The hybrid model allows both to be learned jointly. In cue-based tasks (often used in Parkinson’s EEG studies), brain responses unfold across multiple stages, such as anticipation, movement planning, execution, and feedback. An hybrid model can first extract spatial-spectral activations related to each stage, and the LSTM can then model their temporal order and dependency, which may differ between PD and controls.

The last baseline model is a TCN network, designed to handle sequential data like EEG by using dilated causal convolutions, which allow the network to capture long-range temporal dependencies without the need for recurrent units like in LSTMs. This is crucial for detecting slow, evolving changes in oscillatory patterns or synchronization that are relevant in PD. Unlike LSTMs, TCNs are fully convolutional, meaning they can process entire sequences in parallel. This leads to significantly faster training and inference times. In clinical settings where time efficiency is valuable or when working with large EEG datasets, TCNs provide a clear advantage in computational cost and scalability. Furthermore, TCNs avoid the vanishing gradient problem that often affects RNNs when dealing with long sequences. Their residual connections and stable receptive fields ensure that gradients propagate effectively through deep networks, making TCNs easier to train on long EEG recordings.

The CNN architecture was composed of three sequential convolution–pooling blocks followed by fully connected layers. The LSTM model consisted of a single recurrent layer applied to time–frequency representations reshaped as $(C, T\times F)$ sequences, enabling temporal pattern learning across EEG channels. The hybrid architecture combined convolutional feature extraction with temporal modeling via LSTM layers. The TCN model is composed of four 1D convolutional layers with a dilation factor of $D = 2^n$ with $n \in \{1,..,4\}$, enabling multi-scale feature extraction.

All DL models were trained for a maximum of 100 epochs using the Adam optimizer with a learning rate of $10^-4$ and a batch size of 32. Early stopping with a patience of 5 epochs was employed to prevent overfitting and reduce unnecessary computation during model training phases.

Model performance was quantified using several standard classification metrics: accuracy, precision, recall, F1-score, and the area under the receiver operating characteristic curve (AUC–ROC), all computed on the held-out test set. In addition to predictive performance, we also recorded the average training and inference times for each model across the five validation folds, providing insight into the computational efficiency and scalability of the different approaches.

\section{Results} \label{sec:results}
This section reports the benchmark results achieved using both traditional machine learning approaches and deep learning techniques.

\begin{figure*}[t]
\centering
\includegraphics[width=0.8\linewidth]{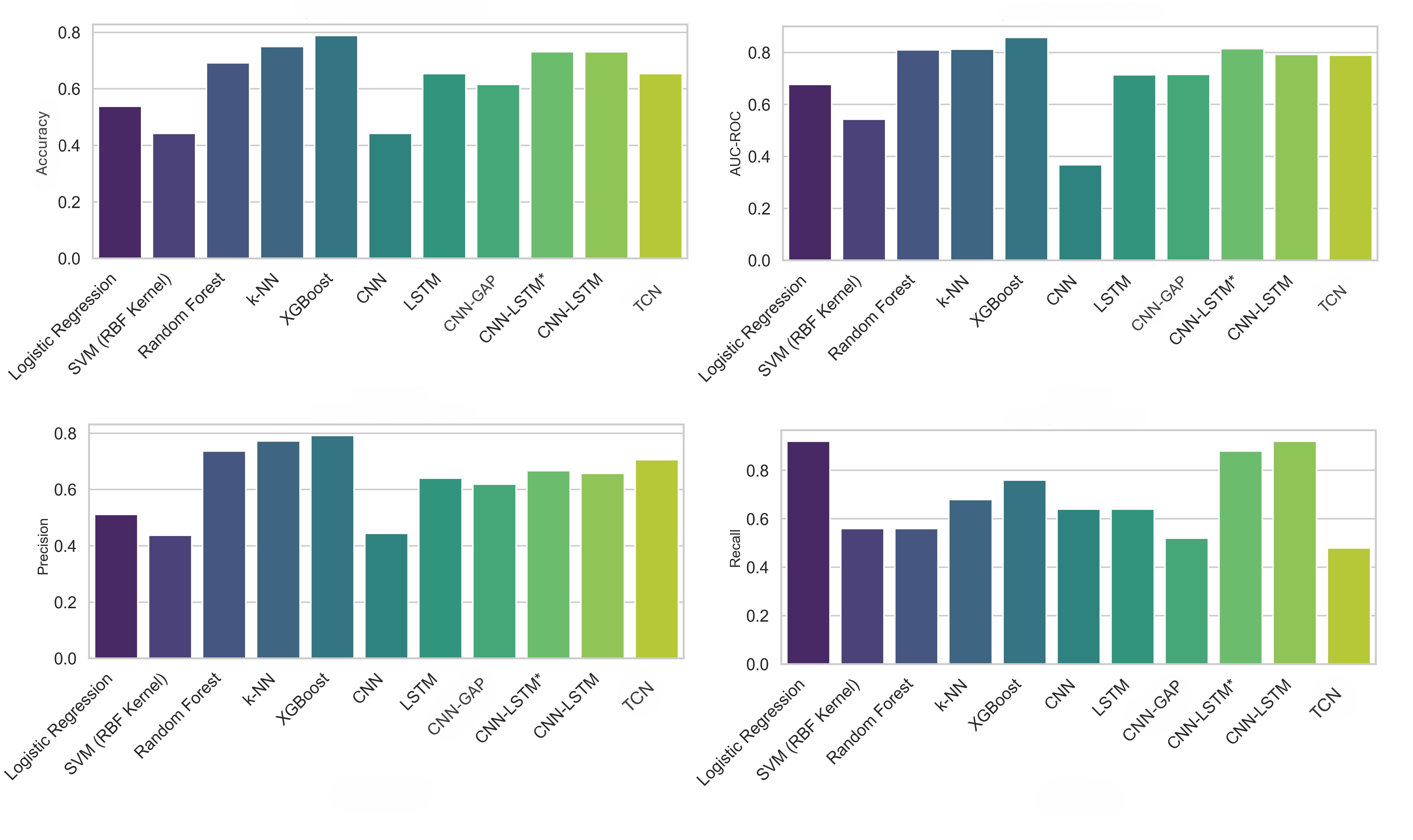}
\caption{Accuracy, AUC-ROC, Precision, and Recall of all eleven classifiers on the held-out test set.}
\label{fig:results}
\end{figure*}

\subsection{Traditional ML Performance}

The results shown in Table~\ref{tab:ml_results} refer to the performance of traditional machine learning models applied to EEG data for the classification of Parkinson’s disease. Among the evaluated classifiers, XGBoost emerged as the most effective model, achieving the highest accuracy (0.79), F1-score (0.78), and AUC–ROC (0.86), indicating that it was able to reliably distinguish between Parkinson’s and control subjects based on EEG-derived features. Its performance also remained consistent in terms of prediction confidence, with a relatively low log loss (0.53).

$k$-NN model followed closely, with an accuracy of 0.75 and a balanced F1-score of 0.72. Its AUC score of 0.81 indicates good discriminative power, although the higher log loss (1.15) suggests that its predicted probabilities were less well-calibrated compared to other models in the study.

The random forest classifier performed moderately, with an accuracy of 0.69 and an F1-score of 0.64. It showed high precision (0.74) but lower recall (0.56), suggesting that it was more conservative and tended to miss a greater number of true Parkinson’s cases. Despite this, it achieved the lowest log loss (0.51), meaning it produced highly confident probability estimates even if not always correct.

In contrast, LR demonstrated an unbalanced profile. While its accuracy was low (0.54), it reached a very high recall (0.92), correctly identifying most Parkinson’s cases but at the cost of many false positives, reflected in the low precision (0.51) and very high log loss (1.96). This suggests a strong bias toward the positive (PD) class, potentially overcompensating for subtle patterns in the data.

SVM achieved the lowest overall performance, with an accuracy of 0.44, an F1-score of 0.49, and an AUC of 0.54. These values indicate that the SVM struggled to capture the relevant structure in the EEG data for PD classification under the selected hyperparameters.

The results confirm that while simpler linear models may struggle with the complexity and variability inherent in EEG data, more flexible approaches such as XGBoost can effectively leverage the spectral-temporal information to identify Parkinson’s disease with good reliability and generalization. Those results can be caused by the EEG-derived features, such as power in different frequency bands across time and channels, that often exhibit nonlinear interactions and complex dependencies that simpler models like logistic regression or SVM with fixed kernels may fail to capture. XGBoost, as a gradient-boosted decision tree ensemble, is particularly well-suited for modeling these kinds of relationships without requiring manual feature engineering or transformations. While models like k-NN may perform well in terms of raw accuracy, XGBoost often delivers more calibrated probability estimates, which contribute to better log loss performance and more reliable clinical decision-making. This is particularly important in healthcare applications where misclassification costs are not symmetric, affecting patient safety outcomes.

\begin{table}[t]
\caption{Traditional ML performance on held-out test set.}
\label{tab:ml_results}
\centering
\begin{tabularx}{\linewidth}{l *{6}{Y}}
\hline
\hline
Model    & Acc.  & Prec. & Rec.  & F1    & AUC   & Log Loss  \\
\hline
LR       & 0.54  & 0.51  & 0.92  & 0.66  & 0.68  & 1.96\\
SVM      & 0.44  & 0.44  & 0.56  & 0.49  & 0.54  & 0.72\\
RF       & 0.69  & 0.74  & 0.56  & 0.64  & 0.81  & 0.51\\
$k$-NN   & 0.75  & 0.77  & 0.68  & 0.72  & 0.81  & 1.15\\
XGB      & 0.79  & 0.79  & 0.76  & 0.78  & 0.86  & 0.53\\
\hline
\hline
\end{tabularx}
\end{table}

\subsection{Deep Learning Performance}

The performance of deep learning models on the held-out EEG test set reveals clear differences in how various architectures handle the temporal and spectral complexity of Parkinson’s-related brain activity (Table~\ref{tab:dl_results}). Among all the evaluated deep models, the two hybrid architectures (Hyb1 and Hyb2), which integrate convolutional and recurrent layers, achieved the best overall performance. Both models reached an accuracy of 0.73, but their strength lay in their ability to balance precision and recall effectively. Hyb2, in particular, showed the highest recall (0.92) and F1-score (0.77), suggesting that it was the most sensitive in detecting Parkinson’s cases, likely due to its capacity to model both localized patterns and long-range dependencies in EEG signals. Hyb1 followed closely, also demonstrating strong recall (0.88) and a high F1-score (0.76), confirming the benefit of combining convolutional feature extraction with sequential modeling for enhanced predictive accuracy.

The LSTM model on its own achieved moderate results, with a balanced accuracy, precision, recall, and F1-score all hovering around 0.64–0.65. This consistency indicates that while the LSTM could capture temporal dynamics present in the EEG sequences, it lacked the representational power of convolutional layers to extract more discriminative features from the raw or preprocessed inputs.

TCN matched the LSTM in terms of accuracy (0.65) and achieved a strong AUC (0.79), yet it produced a noticeably lower recall (0.48), which led to a reduced F1-score (0.57). This suggests that while the TCN was better at making confident distinctions (reflected in its relatively low log loss of 0.54), it tended to be conservative, likely favoring precision over sensitivity to true positives.

Interestingly, the CNN model without global average pooling (GAP) performed the worst among all deep learning models, with an accuracy of 0.44 and the lowest AUC (0.37), indicating poor separability between the two classes. Its recall was relatively high (0.64), but the low precision (0.44) and F1-score (0.52) reflect a tendency toward over-predicting the Parkinson’s class. This may point to overfitting on local features or a lack of global context in the decision-making process.

Adding global average pooling (CNN-GAP) significantly improved performance over the standard CNN, raising accuracy to 0.62 and AUC to 0.72. Although recall dropped slightly (0.52), both precision (0.62) and F1-score (0.57) improved, suggesting that global average pooling helped stabilize predictions and made the model more discriminative overall. The improvement highlights the value of global abstraction in EEG classification tasks, particularly when spatial or frequency-related patterns are distributed rather than sharply localized.

As depicted in Figure~\ref{fig:results}), these results suggest that while simple CNNs struggle with the temporal structure of EEG data, architectures that combine temporal modeling (such as LSTMs) with spatial feature extraction (via CNNs) are better equipped to capture the pathological dynamics of Parkinson’s disease. The hybrid models, in particular, demonstrate a strong ability to generalize, offering the best trade-off between sensitivity and precision. The performance of the TCN also confirms the relevance of sequence modeling, although its conservative behavior might require further tuning or integration with additional layers to improve recall. Overall, these findings reinforce the importance of modeling both temporal evolution and local feature patterns when designing deep learning architectures for EEG-based diagnosis in clinical decision systems..

\begin{table}[t]
\caption{Deep learning performance on held-out test set}
\label{tab:dl_results}
\centering
\begin{tabularx}{\linewidth}{l *{6}{Y}}
\hline
\hline
Model       & Acc.  & Prec. & Rec.  & F1    & AUC   & Log Loss \\
\hline
CNN         & 0.44  & 0.44  & 0.64  & 0.52  & 0.37  & 0.71\\
CNN-GAP        & 0.62  & 0.62  & 0.52  & 0.57  & 0.72  & 0.61\\
LSTM        & 0.65  & 0.64  & 0.64  & 0.64  & 0.72  & 0.63\\
Hyb1        & 0.73  & 0.67  & 0.88  & 0.76  & 0.81  & 0.56\\
Hyb2        & 0.73  & 0.66  & 0.92  & 0.77  & 0.79  & 0.56\\
TCN         & 0.65  & 0.71  & 0.48  & 0.57  & 0.79  & 0.54\\
\hline
\hline
\end{tabularx}
\end{table}

\section{Conclusion} \label{sec:conclusion}

In this study, we conducted a comprehensive benchmark of traditional machine learning and deep learning baseline models for the classification of Parkinson’s disease using EEG recordings from a well-structured oddball task dataset. Our aim was to systematically evaluate a wide range of baseline models within a unified framework, enabling fair comparisons across different learning paradigms and shedding light on the effectiveness of temporal, spatial, and spectral features in neurophysiological diagnosis.

The study demonstrated that classical ML approaches, particularly XGBoost and $k$-NN, can achieve strong performance when applied to carefully engineered spectral features. XGBoost, in particular, exhibited robust classification accuracy, high AUC, and low log loss, confirming its suitability for clinical tasks involving EEG-derived measures. These results underscore the continued relevance of handcrafted feature-based approaches, especially when paired with powerful ensemble learning techniques.

However, our results also highlighted the unique strengths of deep learning models in capturing the complex temporal and spatial dynamics inherent in EEG data. Among the deep models evaluated, hybrid CNN–LSTM architectures delivered the most consistent and high-performing results, effectively integrating local spatial patterns with temporal sequence modeling. This suggests that the pathophysiological alterations associated with Parkinson’s disease manifest not only in frequency content but also in how brain activity evolves across time. The performance of the TCN model, though slightly more conservative in recall, further reinforces the importance of capturing long-range temporal dependencies in EEG signals.
Importantly, our use of a unified preprocessing pipeline, strict subject-wise data splitting, and comprehensive evaluation metrics ensured a rigorous and unbiased comparison. This methodology helps clarify the relative merits of ML and DL strategies, which is critical given the inconsistencies and fragmented methodologies often found in existing literature.
Future research could extend this work by incorporating additional task paradigms, exploring subject-specific adaptation techniques, and increasing the dataset dimension to enhance generalization and study the performance of ML algorithms compared to DL models in these scenarios. 

In conclusion, our findings demonstrate that both traditional and deep learning models can effectively support EEG-based diagnosis of Parkinson’s disease, even with a limited amount of data, an often encountered scenario in medical research. Furthermore, we have demonstrated the effectiveness of capturing long-range temporal dependencies using baseline methods, which will also be applicable when employing more complex models. Hybrid architectures present a promising path forward in this area. This benchmark provides a solid foundation for future work and a valuable reference point for the design of next-generation neurodiagnostic tools.

\section*{Acknowledgments}
This work was supported by ``Smart unmannEd AeRial vehiCles for Human likE monitoRing (SEARCHER)'' project of the Italian Ministry of Defence within the PNRM 2020 Program (Grant Number: PNRM a2020.231); ``EYE-FI.AI: going bEYond computEr vision paradigm using wi-FI signals in AI systems'' project of the Italian Ministry of Universities and Research (MUR) within the PRIN 2022 Program (Grant Number: 2022AL45R2) (CUP: B53D23012950001); MICS (Made in Italy – Circular and Sustainable) Extended Partnership and received funding from Next-Generation EU (Italian PNRR – M4 C2, Invest 1.3 – D.D. 1551.11-10-2022, PE00000004) (CUP MICS B53C22004130001); and “Enhancing Robotics with Human Attention Mechanism via Brain-Computer Interfaces” Sapienza University Research Projects (Grant Number: RM124190D66C576E).

\balance
\bibliographystyle{latex8}
\bibliography{bibliography}

\end{document}